\documentclass[10pt]{article}
\usepackage[preprint]{tmlr}

\usepackage{amsmath,amssymb}
\usepackage{booktabs}
\usepackage{array}
\usepackage{graphicx}
\usepackage{multirow}
\usepackage{xcolor}
\usepackage{microtype}
\usepackage{hyperref}
\usepackage{url}
\usepackage{xspace}
\usepackage{enumitem}
\setlist{itemsep=2pt,parsep=0pt,topsep=4pt}
\definecolor{ours}{HTML}{2A78D6}
\hypersetup{colorlinks=true,linkcolor=ours,citecolor=ours,urlcolor=ours}

\newcommand{\bfs}{bf16\xspace}
\newcommand{\RTN}{RTN\xspace}
\newcommand{\SR}{SR\xspace}
\newcommand{\Flog}{F_{\log}}

\title{Lost in Rounding: bf16 Recurrent State Makes\\Hybrid Reasoning Models Think Longer}
\author{\name Yuchuan Weng \email yuweng@paypal.com \\
        \name Xin Chen \email xchen7@paypal.com \\
        \name Chi Liu \email chiliu@paypal.com \\
        \addr PayPal AI}

\def\month{11}
\def\year{2026}
\def\openreview{}

\begin{document}
\maketitle

\begin{abstract}
Hybrid reasoning models interleave attention with recurrent mixers such as Mamba-2 and Gated DeltaNet, whose state is stored in a cache between decoding steps.
Prior work showed that storing this state in FP16 or INT8 with round-to-nearest (\RTN) makes Mamba-2 and GDN hybrids more verbose, and that stochastic rounding (\SR) fixes it.
We study the other 16-bit format that serving engines expose, \bfs, on six open hybrid reasoning models and 590 competition-math problems.
With \bfs \RTN state, five of the six models reason longer (geometric-mean length $\times$1.07 to $\times$1.27; also under greedy decoding in the two models tested) and hit the 32K token cap more often.
On Qwen3.5-9B, cap hits rise from 118 to 223 and accuracy falls from 85.6\% to 70.8\%.
On LiveCodeBench, both models tested also reason longer and lose 12.6 and 13.4 points of pass@1.
Rerunning the capped items with a 128K budget halves the accuracy gap but does not close it (93.6\% vs.\ 86.6\%): \bfs runs that never finish are twice as common, and 85\% of them are exact repetition loops.
On Nemotron-Nano-9B-v2, by contrast, 30 of 31 extra caps finish at 128K, so the cost there is mostly extra tokens (exploratory).
Same-width \bfs \SR removes the length increase in four of the five affected models, restores Qwen3.5-9B to 87.6\% and brings code length and pass@1 near FP32 levels, so the failure is mostly rounding bias, not missing resolution.
With an integer-hash random generator, \SR on Qwen3.5-9B matches FP32 on cap hits and accuracy (113 vs.\ 118; 85.6\% for both) at the memory and decoding speed of \bfs \RTN.
In the Nemotron-Nano-v2 lineage, \SR on a band of intermediate-decay heads (25\% of heads) is enough to remove the effect, while random heads and the two decay octiles with the largest teacher-forced KL share are not; in Nemotron-3 and Qwen3.5-9B the effect is instead spread across heads.
We recommend that evaluations of hybrid models report the state dtype, cap rate and output length alongside accuracy.
\end{abstract}

\section{Introduction}

Hybrid models combine a few attention layers with many recurrent layers, such as Mamba-2~\citep{mamba2} or Gated DeltaNet (GDN)~\citep{gdn}.
At decode time each recurrent layer reads its state from a cache, updates it once per token, and writes it back.
The state is therefore re-rounded at every step, so any rounding bias can accumulate over a long reasoning trace.
A \bfs cache takes half the memory of FP32 and, on our servers, admits twice as many requests per GPU (Appendix~\ref{app:serving}).

Lengthening from rounding bias is already documented for some formats.
Nemotron~3 Super reports that casting the SSM cache to FP16 increases verbosity by up to 37\%, attributes this to round-to-nearest bias that ``accumulates coherently over time'', and fixes it with stochastic rounding~\citep{super}.
Nemotron~3 Ultra extends the finding to INT8 and FP8 caches~\citep{ultra}.
\citet{lores} find the same for INT8 state on Qwen-3.5 and Nemotron-H: \RTN lengthens responses and exhausts the budget more often, while \SR stays close to the FP32 baseline.
For \bfs, the evidence is mixed.
A vLLM bug fix reports that \bfs rounding of the recurrent state ``shows up as runaway generations'' in one Nemotron model and changes the default to FP32~\citep{vllm63218}. In that model, FP32 cut generations hitting the token cap from 23\% to 0.25\% and raised requests/s 2.55$\times$.
LeapQuant reports that \bfs state lowers Qwen3.5-9B's AIME score from 87.9\% to 72.1\%, but finds that \SR ``helps little'' for state error~\citep[\S\S3.2, 4.2]{leapquant}.
DAMP finds that FP16 beats \bfs at equal storage and attributes this to FP16's finer resolution~\citep{damp}.
In SGLang, \SR is available only for FP16 Mamba caches~\citep{sgl26929}; a pull request adding \bfs \SR for GDN and Mamba was closed without accuracy results~\citep{sgl27975}.

We build on this work with a controlled study of \bfs \RTN state on six open hybrid reasoning models (two related Mamba-2 models, a Mamba-2 MoE, two GDN models and a parallel hybrid).
Our contributions are:
\begin{enumerate}
\item \textbf{A paired six-model measurement at \bfs.} Against FP32 on the same prompts, \bfs \RTN lengthens reasoning in five of six models and raises cap hits, under both sampling and greedy decoding (\S\ref{sec:effect}), and on code in both models tested (\S\ref{sec:lcb}).
\item \textbf{Budget versus non-termination.} Rerunning capped items at 128K shows that about half of Qwen3.5-9B's accuracy loss disappears with more budget, while the rest comes from runs that never finish, mostly exact loops; on Nemotron-Nano-9B-v2 nearly all extra caps finish (\S\ref{sec:budget}). The common ``both-finished'' accuracy split~\citep{replayssm} hides this.
\item \textbf{Bias, not resolution.} Same-width \bfs \SR, which we implement in SGLang's Mamba-2 and GDN decode kernels, removes the effect in four of five affected models (\S\ref{sec:sr}). This extends the FP16 and INT8 results of \citet{super,ultra,lores} to \bfs, on both Mamba-2 and GDN models, and argues against a resolution-only explanation of the \bfs gap~\citep{damp}.
\item \textbf{\bfs \SR at \bfs cost.} \SR writes to the same \bfs cache as \RTN, so it needs no extra memory, and its only cost is random-bit generation. With Triton's default Philox our GDN kernel decodes at 30--60\% of \RTN speed; with an integer hash, \SR on Qwen3.5-9B matches the throughput of a same-day \RTN run, end to end and at matched batch size, and FP32's length, cap hits, loops and accuracy (pre-registered; \S\ref{sec:sr}).
\item \textbf{Teacher-forced KL misranks head groups.} In Nemotron-Nano-v2, \SR on a band of intermediate-decay heads (two decay octiles, 25\% of heads) removes essentially all of the effect, while random heads and the two octiles with the largest teacher-forced KL share do not (\S\ref{sec:heads}). In Nemotron-3 and Qwen3.5-9B the effect is spread across heads, so head selection has to be validated per model.
\end{enumerate}
Key tests were pre-registered: we wrote predictions and pass/fail criteria in a working notebook before running each arm.
The notebook is not externally timestamped; Appendix~\ref{app:prereg} lists the criteria.
We label the remaining analyses exploratory.

\section{Background}
\label{sec:background}

\begin{figure}[t]
\centering
\includegraphics[width=0.64\linewidth]{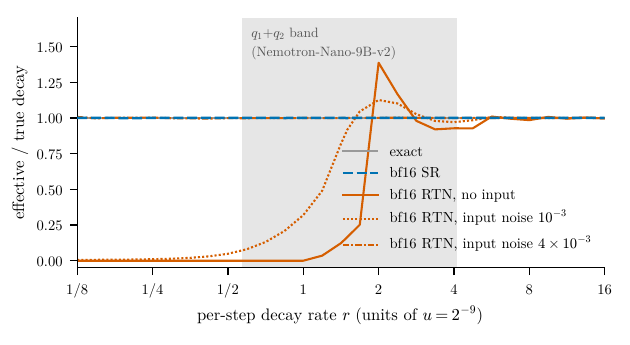}
\caption{Effective per-step decay of a quiet leaky integrator $h \leftarrow \mathrm{round}(e^{-r} h + \epsilon)$ stored in \bfs, relative to the true rate, over 20{,}000 trajectories. Input noise $\epsilon$ has the given standard deviation relative to the initial state. \RTN freezes slow decay and distorts it up to about $4u$; \SR is unbiased at every rate. The shaded band is the decay range of the Nemotron-Nano-9B-v2 heads whose \SR removes the length effect (\S\ref{sec:heads}); the toy explains its upper edge but not its lower edge.}
\label{fig:toy}
\end{figure}

A Mamba-2 head updates $h_t = e^{-\Delta_t a}\,h_{t-1} + \Delta_t B_t x_t$, with $a>0$ and step size $\Delta_t = \mathrm{softplus}(\cdot + b_\Delta)$; a GDN head updates a matrix state with a gated delta rule~\citep{gdn}.
Serving engines compute the update in FP32 and store the result in the cache dtype.
We call $r = e^{A_{\log}}\cdot\mathrm{softplus}(b_\Delta)$ a head's \emph{weights-only decay rate}, the per-step decay when the input-dependent part of $\Delta_t$ is zero.

bf16 keeps 8 significant bits.
Let $u = 2^{-9}$ be its smallest relative half-spacing.
If a head's per-step change is below about $u$ relative to the stored value, \RTN rounds the update away and the state does not move.
This ``write erasure'' was formalised by \citet{dither}, and \citet{preserving} show that such lost small updates make state quantization errors coherent across steps.
A one-line simulation of a quiet leaky integrator $h \leftarrow \mathrm{round}(e^{-r} h)$ illustrates the regimes: \RTN freezes decay completely for $r \lesssim u$, distorts it for $u < r \lesssim 4u$ (effective rate $0.17\times$ at $1.5u$, $1.39\times$ at $2u$), and is nearly unbiased above.
\SR rounds up with probability equal to the fractional position between the two neighbouring \bfs values~\citep{gupta}, so it is unbiased at every rate.
We use this only as an illustration (Figure~\ref{fig:toy}); it does not predict which heads matter (\S\ref{sec:heads}).

\section{Setup}
\label{sec:setup}

\paragraph{Models.}
Nemotron-Nano-9B-v2 and its parent Nemotron-Nano-12B-v2~\citep{nano2} (Mamba-2 hybrids; one lineage);
Nemotron-3-Nano-30B-A3B~\citep{nemo3nano} (Mamba-2 MoE hybrid);
Qwen3.5-9B~\citep{qwen35} and Qwen3-Next-80B-A3B-Thinking~\citep{qwen3next} (GDN hybrids);
Falcon-H1R-7B~\citep{falconh1r,falconh1} (attention and Mamba-2 in parallel in every layer).
All weights and activations are bf16; only the recurrent state dtype changes between arms.

\paragraph{Prompts and decoding.}
590 problems: MATH-500~\citep{math,prm800k} plus AIME 2024, AIME 2025 and HMMT February 2025 (30 each; MathArena release); for two models, also 175 LiveCodeBench problems (\S\ref{sec:lcb}).
We sample with each model card's settings: $T{=}0.6$, top-$p$ 0.95 (Qwen models also top-$k$ 20), except Nemotron-3 at $T{=}1.0$, top-$p$ 1.0.
Greedy arms use $T{=}0$.
The output cap is 32{,}768 new tokens unless stated.

\paragraph{Serving.}
SGLang 0.5.21~\citep{sglang} on 8 RTX PRO 6000 Blackwell GPUs, with \texttt{-{}-mamba-ssm-dtype} set to \texttt{float32}, \texttt{bfloat16} or \texttt{float16} (three FP32 arms used the \texttt{float32} default; Appendix~\ref{app:serving}).
Most arms use 8-way data parallelism; Qwen3-Next uses TP4$\times$DP2.
The Qwen3.5-9B FP32 arm ran with 6 data-parallel ranks; an 8-rank rerun agrees with it (accuracy 86.6\% vs.\ 85.6\%, McNemar $p{=}0.58$; length ratio between the two FP32 runs $\times$1.017).
Because an FP32 state doubles cache size, \bfs arms can admit twice as many concurrent requests; we control for this on Nemotron-3 (\S\ref{sec:effect}), and on Qwen3.5-9B the cap never binds (Appendix~\ref{app:serving}).
Each arm is one seed unless stated.

\paragraph{Stochastic rounding.}
We patch SGLang's Triton selective-state-update kernel (Mamba-2) and its packed GDN decode kernel.
When enabled, the kernel adds 16 random bits below the \bfs mantissa of each FP32 state value and then truncates, so the value is rounded up with probability equal to its fractional position between the two \bfs neighbours (at 16-bit resolution). Random bits come from Triton's counter-based generator, with a fresh seed at every CUDA-graph replay.
Storage is identical to \bfs \RTN, and so are the engine's concurrency limits: the server logs show the same per-rank request cap and cache slot count for the \SR and \RTN arms of all five models.
With \SR disabled the patched kernels are bitwise identical to the originals.
A per-head mask applies \SR to listed heads and \RTN to the rest.
Prefill state writes still use \RTN, once per request in our runs (Appendix~\ref{app:serving}).

\paragraph{Metrics.}
\emph{Length ratio}: geometric mean of paired per-item output-length ratios, with a Wilcoxon signed-rank test on log ratios.
\emph{Capped}: the run stopped at the token cap.
\emph{Accuracy}: the last \texttt{\textbackslash boxed\{\}} answer in the output, checked with \texttt{math-verify}; capped runs are graded the same way, so a capped run whose last boxed answer is correct counts as correct; paired arms are compared with an exact McNemar test.
\emph{Excess removed} by an intervention: $\Flog = (\overline{\log L}_{\mathrm{bf16}} - \overline{\log L}_{\mathrm{arm}})/(\overline{\log L}_{\mathrm{bf16}} - \overline{\log L}_{\mathrm{fp32}})$, with a 95\% paired-bootstrap interval (4{,}000 resamples).
$\Flog{=}1$ means the arm matches FP32 length; $\Flog{=}0$ means it matches \bfs \RTN.

\section{bf16 RTN state lengthens reasoning}
\label{sec:effect}

\begin{figure}[t]
\centering
\includegraphics[width=\linewidth]{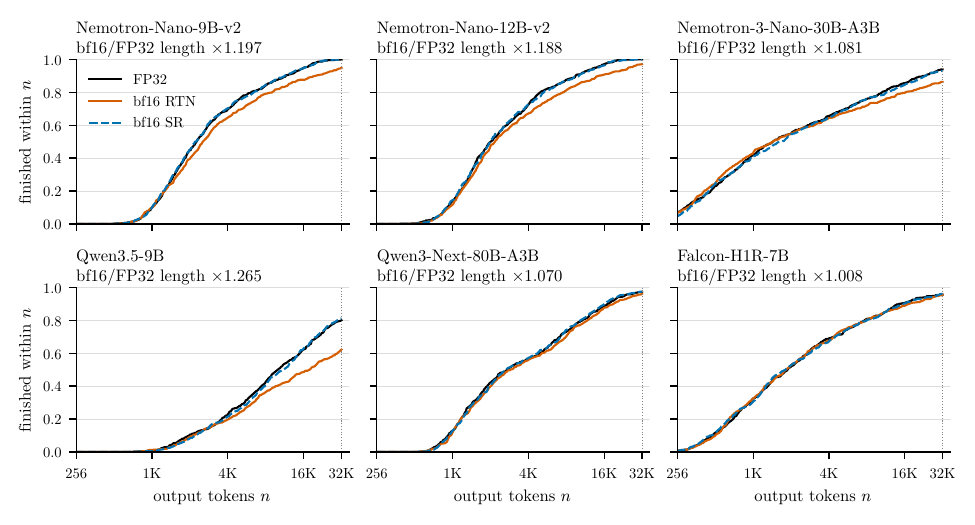}
\caption{Share of the 590 items finished within $n$ output tokens; items that hit the 32K cap never finish, so each curve ends at one minus the cap rate. Titles give the \bfs/FP32 length ratio (Table~\ref{tab:effect}).}
\label{fig:completion}
\end{figure}

\begin{table}[t]
\centering
\small
\caption{FP32 vs.\ \bfs \RTN recurrent state ($n{=}590$, 32K cap). Length ratio is \bfs/FP32. ``Both finished'' is accuracy on items that neither arm capped.}
\label{tab:effect}
\footnotesize\setlength{\tabcolsep}{4pt}
\begin{tabular}{@{}lcccccc@{}}
\toprule
Model (mixer) & Length ratio & Capped & Accuracy (\%) & McNemar & \multicolumn{2}{c}{Both finished} \\
 & ($p$) & FP32$\to$\bfs & FP32$\to$\bfs & $p$ & $n$ & FP32 / \bfs \\
\midrule
\multicolumn{7}{@{}l}{\emph{Sampling}}\\
Nemotron-Nano-9B-v2 (M2) & $\times$1.197 (7e-19) & 1$\to$31 & 89.3$\to$88.8 & 0.74 & 558 & 92.7 / 93.5 \\
Nemotron-Nano-12B-v2 (M2) & $\times$1.188 (1e-18) & 0$\to$17 & 90.0$\to$90.5 & 0.70 & 573 & 92.1 / 93.2 \\
Nemotron-3-Nano-30B-A3B (M2) & $\times$1.081 (3e-6) & 36$\to$81 & 92.5$\to$87.3 & 3e-7 & 509 & 98.4 / 98.6 \\
Qwen3.5-9B (GDN) & $\times$1.265 (7e-16) & 118$\to$223 & 85.6$\to$70.8 & 2e-14 & 322 & 99.7 / 99.7 \\
Qwen3-Next-80B-A3B (GDN) & $\times$1.070 (5e-12) & 16$\to$24 & 95.9$\to$94.6 & 0.10 & 561 & 99.1 / 99.1 \\
Falcon-H1R-7B (parallel M2) & $\times$1.008 (0.66) & 26$\to$27 & 92.9$\to$93.4 & 0.70 & 553 & 97.3 / 97.8 \\
\midrule
\multicolumn{7}{@{}l}{\emph{Greedy}}\\
Nemotron-Nano-9B-v2 & $\times$1.211 (6e-25) & 5$\to$49 & 89.3$\to$87.5 & 0.13 & & \\
Qwen3.5-9B & $\times$1.172 (6e-10) & 208$\to$285 & 75.6$\to$59.8 & 1e-11 & & \\
\bottomrule
\end{tabular}
\end{table}

Table~\ref{tab:effect} shows the main result.
Five of six models produce longer outputs with \bfs \RTN state, and cap hits rise in all five, significantly in four (binomial test on discordant caps; Qwen3-Next 16$\to$24, $p{=}0.10$).
Figure~\ref{fig:completion} shows the whole length distribution: \bfs \RTN shifts it to the right and leaves more items unfinished at the cap, while \SR (\S\ref{sec:sr}) tracks FP32 except on Nemotron-3.
Falcon-H1R-7B shows no effect; we do not have an explanation.
The effect is also present under greedy decoding (pre-registered; both models pass), so it is not a sampling artifact.
It holds across seeds: a second Nemotron-Nano-9B-v2 seed gives $\times$1.231 with caps 0$\to$32.

On Nemotron-Nano-9B-v2 the increase concentrates on harder problems: $\times$1.15 on MATH-500 but $\times$1.48, $\times$1.40 and $\times$1.68 on AIME24, AIME25 and HMMT.
Qwen3.5-9B does not show this pattern ($\times$1.27, $\times$1.32, $\times$1.31 and $\times$1.16).
FP16 \RTN on Nemotron-Nano-9B-v2 is nearly neutral overall ($\times$1.017, $p{=}0.13$) but reaches $\times$1.25 on HMMT (30 problems; exploratory), consistent with the FP16 verbosity reported for longer generations by \citet{super}.

Two controls address confounds on Nemotron-3.
Matching the \bfs arm's per-rank concurrency to the FP32 arm (43 requests) leaves the effect intact ($\times$1.098, $p{=}4\times10^{-6}$; caps 82; pre-registered).
A second FP32 seed agrees with the first ($\times$0.982, $p{=}0.56$), so the effect is not seed noise (exploratory).

Accuracy drops significantly for Nemotron-3 and Qwen3.5-9B, and in both cases almost all lost items are \bfs runs that hit the cap: 32 of 35 for Nemotron-3, and all 112 for Qwen3.5-9B.
Accuracy on items that both arms finish is unchanged.
This subset is selected, however: it contains the items short enough to finish under both settings, and accuracy on it is near ceiling (99.7\% for Qwen3.5-9B).
It cannot show that \bfs costs nothing beyond the budget.
The next section tests that directly.

\section{More budget: longer reasoning versus non-termination}
\label{sec:budget}

\begin{table}[t]
\centering
\small
\caption{Items capped at 32K in either arm, rerun at 128K with the same sampling settings. Qwen3.5-9B is pre-registered; Nemotron-Nano-9B-v2 (32 items) is exploratory. ``Relaxed'' accuracy covers all 590 items, using the 128K result for rerun items and the 32K result otherwise.}
\label{tab:128k}
\begin{tabular}{@{}lcccc@{}}
\toprule
 & \multicolumn{2}{c}{Qwen3.5-9B ($n{=}268$)} & \multicolumn{2}{c}{Nemotron-Nano-9B-v2 ($n{=}32$)} \\
\cmidrule(lr){2-3}\cmidrule(l){4-5}
 & FP32 & \bfs \RTN & FP32 & \bfs \RTN \\
\midrule
Unfinished at 128K & 44 & 93 & 0 & 1 \\
\quad of which exact repetition loops & 42 & 79 & 0 & 1 \\
\bfs-only 32K caps that finish at 128K & & 108 / 150 & & 30 / 31 \\
Accuracy among finished reruns & 97.3\% & 97.7\% & 37.5\% & 25.8\% \\
Median output length on rerun items & 19{,}951 & 41{,}949 & 16{,}314 & 37{,}401 \\
Relaxed accuracy, all 590 items & 93.6\% & 86.6\% & 89.7\% & 89.8\% \\
\bottomrule
\end{tabular}
\end{table}

We reran every Qwen3.5-9B item that was capped in either arm (268 items) at 128K, with the same sampling settings (Table~\ref{tab:128k}).
Reruns sample from scratch rather than continue the 32K runs, identically in both arms, so they ask whether an item can finish, not how a given run would have continued.
Both pre-registered tests pass.
The relaxed accuracy gap shrinks from 14.7 points at 32K to 6.9 points (41 of 590 items), but stays significant (55 vs.\ 14 discordant items, $p{=}7\times10^{-7}$).
\bfs runs are more often unfinished even at 128K (17 FP32-only vs.\ 66 \bfs-only, $p{=}6\times10^{-8}$).
A run counts as an exact loop when its tail repeats with a fixed period for at least two periods and 192 tokens; 79 of the 93 unfinished \bfs runs (85\%) meet this.
At 32K, exact loops among capped runs already rose from 72 to 113 (45 vs.\ 86 discordant items, $p{=}4\times10^{-4}$); in the other five models they are rare in both arms (at most two per arm), so looping is a Qwen3.5-9B result.

On the 158 rerun items that both arms finish at 128K, accuracy is 99.4\% for FP32 and 97.5\% for \bfs (3 vs.\ 0 discordant, $p{=}0.25$); this subset is again selected and near ceiling.
So \bfs \RTN causes two failures: longer reasoning that finishes when given more budget, and runs that do not terminate at any tested budget.
Weight quantization shows a related failure: in up to 52\% of quantized models' failures, the model reaches the right answer in intermediate steps but does not output it as the final answer~\citep{lotfi}.
Here the cause is the state precision, and many such runs end in exact loops.
Of the 150 items capped at 32K only under \bfs, 108 finish at 128K and 42 do not, 40 of them in exact loops.
The second failure is invisible in a both-finished split~\citep{replayssm} and in a truncation-waste decomposition~\citep{couplingtax}, which both condition on the output budget.

The mix of the two failures depends on the model.
In an exploratory rerun of the 32 Nemotron-Nano-9B-v2 items capped in either arm, 30 of the 31 \bfs-only caps finish at 128K, and only one run loops.
\bfs reasoning on these items is $\times$2.29 longer (geometric mean), but relaxed accuracy is the same (89.7\% vs.\ 89.8\%).
On the 32 items alone FP32 solves 12 and \bfs 8 (7 vs.\ 3 discordant, $p{=}0.34$), too few to separate.
For this model the cost of \bfs \RTN is mainly extra tokens; for Qwen3.5-9B it also includes runs that never end.

\section{Same-width stochastic rounding removes the effect}
\label{sec:sr}

\begin{table}[t]
\centering
\small
\caption{\bfs \SR on all heads, same storage as \bfs \RTN. $\Flog$ is the share of the \bfs log-length excess removed. Length ratio is relative to FP32. ``Hash generator'': an integer hash instead of Triton's Philox.}
\label{tab:sr}
\footnotesize\setlength{\tabcolsep}{3.8pt}
\begin{tabular}{@{}lccccc@{}}
\toprule
Model & $\Flog$ [95\% CI] & Length ratio & Capped & Accuracy (\%) & vs.\ FP32 \\
 & & \SR/FP32 & FP32 / \RTN{} / \SR & FP32 / \RTN{} / \SR & McNemar $p$ \\
\midrule
Nemotron-Nano-9B-v2 & 1.05 [0.88, 1.27] & $\times$0.991 & 1 / 31 / 0 & 89.3 / 88.8 / 89.7 & 0.88 \\
Nemotron-Nano-12B-v2 & 0.98 [0.82, 1.16] & $\times$1.004 & 0 / 17 / 0 & 90.0 / 90.5 / 90.3 & 0.86 \\
Nemotron-3-Nano-30B-A3B & 0.38 [$-$0.37, 1.00] & $\times$1.049 & 36 / 81 / 41 & 92.5 / 87.3 / 92.2 & 0.81 \\
Qwen3.5-9B & 0.87 [0.58, 1.26] & $\times$1.031 & 118 / 223 / 111 & 85.6 / 70.8 / 87.6 & 0.24 \\
\quad hash generator & 0.97 [0.70, 1.37] & $\times$1.007 & 118 / 223 / 113 & 85.6 / 70.8 / 85.6 & 1.00 \\
Qwen3-Next-80B-A3B & 0.99 [0.68, 1.41] & $\times$1.001 & 16 / 24 / 15 & 95.9 / 94.6 / 96.1 & 1.00 \\
\bottomrule
\end{tabular}
\end{table}

If the \bfs failure came from too little resolution, an unbiased rounding at the same width should not help much.
Table~\ref{tab:sr} shows that it does.
In four of five affected models, \SR returns length, cap rate and accuracy to FP32 levels.
On Qwen3.5-9B it recovers 121 items that \bfs \RTN lost and loses 22, all of the former capped under \RTN and all of the latter capped under \SR.
The Nemotron-Nano-12B-v2, Qwen3.5-9B and Qwen3-Next tests were pre-registered and pass.

Nemotron-3 is the exception.
Its pre-registered test fails: $\Flog = 0.38$.
\SR does remove the runaway tail (40 of the 45 extra caps; accuracy 92.2\% vs.\ 92.5\% for FP32), but a bulk shift of about $\times$1.05 remains, and it also appears when comparing against a second FP32 seed.
We do not know its cause.

Our GDN \SR arms use Triton's default Philox generator, which makes that kernel slow (\S\ref{sec:limits}).
A pre-registered rerun of Qwen3.5-9B with the integer hash lowbias32~\citep{lowbias32} instead removes the effect as fully: $\Flog = 0.97$ [0.70, 1.37]; 113 caps, 73 of them exact loops (FP32: 118 and 72; \bfs \RTN: 223 and 113); accuracy 85.6\%, equal to FP32.
It ran at 10{,}393 tokens/s end to end, against 10{,}163 for a same-day \bfs \RTN run, which itself reproduces the effect (230 caps, 70.0\%); at matched batch size in the server logs, measured as for the Mamba-2 arms (\S\ref{sec:limits}), it decodes no slower than that run (102--113\% across batch bins).
So on this model \SR removes the effect at \bfs cost in both memory and speed.

These results extend the FP16 and INT8 findings of \citet{super,ultra,lores} to \bfs on both mixer types; \citet{lores} already covered GDN at INT8.
They also refine two \bfs results.
LeapQuant finds that \SR ``helps little'' for \bfs state error, lowering it only $3\times$ at 64K tokens~\citep[\S3.2]{leapquant}; we find that \SR nonetheless removes the behavioural effect, which is consistent with the effect being caused by bias rather than error magnitude.
DAMP attributes FP16's advantage over \bfs to resolution~\citep{damp}; at the same \bfs resolution, removing the bias is enough in most of our models.
Resolution still matters for \RTN: FP16 \RTN on Nemotron-Nano-9B-v2 is much milder than \bfs \RTN (\S\ref{sec:effect}).

\section{Code generation}
\label{sec:lcb}

\begin{table}[t]
\centering
\small
\caption{LiveCodeBench v6 increment, 175 problems $\times$ 2 samples, 32K cap. Length ratio is \bfs \RTN/FP32 with a 95\% bootstrap CI over problems; $\Flog$ as in Table~\ref{tab:sr}. Pass@1 is exploratory.}
\label{tab:lcb}
\footnotesize\setlength{\tabcolsep}{3.5pt}
\begin{tabular}{@{}lcccc@{}}
\toprule
Model & Length ratio & $\Flog$ [95\% CI] & Capped (of 350) & Pass@1 (\%) \\
 & \bfs/FP32 [95\% CI] & & FP32 / \RTN{} / \SR & FP32 / \RTN{} / \SR \\
\midrule
Nemotron-Nano-9B-v2 & $\times$1.37 [1.29, 1.46] & 0.95 [0.77, 1.15] & 36 / 162 / 30 & 58.3 / 45.7 / 59.4 \\
Qwen3.5-9B & $\times$1.09 [1.05, 1.13] & 0.73 [0.27, 1.33] & 169 / 233 / 179 & 46.3 / 32.9 / 45.4 \\
\bottomrule
\end{tabular}
\end{table}

To test whether the effect is specific to math, we ran Nemotron-Nano-9B-v2 and Qwen3.5-9B on the 175 problems of the LiveCodeBench~\citep{lcb} v6 increment (contest dates 4 January to 6 April 2025), with two samples per problem and the settings of \S\ref{sec:setup}.
The prompt is the benchmark's generic code-generation prompt.
A sample passes if the last code block after its reasoning passes all public and private tests under the official harness.
The unit of analysis is the problem: we average log length over its two samples and bootstrap over problems.
The length, \SR and cap tests were pre-registered (Appendix~\ref{app:prereg}).

Both models reproduce the math result (Table~\ref{tab:lcb}).
\bfs \RTN lengthens outputs, more problems gain cap hits than lose them (65 vs.\ 2 for Nemotron-Nano-9B-v2, 39 vs.\ 3 for Qwen3.5-9B; sign test $p < 10^{-8}$), and \SR removes most of the excess.
On Nemotron-Nano-9B-v2 the length effect is larger than on math ($\times$1.37 vs.\ $\times$1.20).
Pass@1 falls by 12.6 points [6.9, 18.3], and \SR restores it.
Qwen3.5-9B already caps 48\% of FP32 samples, which compresses its ratio toward 1; on the 86 problems where no FP32 sample capped, the ratio is $\times$1.20 (exploratory; we committed in advance to report it if the FP32 cap rate exceeded 30\%).
Its pass@1 falls by 13.4 points [8.3, 19.1], and \SR recovers 12.6 of them; its $\Flog$ interval is wide but passes the pre-registered test.

Pass@1 among samples that finish is higher under \bfs \RTN (85.1\% vs.\ 65.0\% for Nemotron-Nano-9B-v2, 98.3\% vs.\ 89.5\% for Qwen3.5-9B).
This is the selection effect of \S\ref{sec:effect}: a finished-only metric would rank \bfs \RTN above FP32.
Exact loops do not explain the extra caps: under \bfs \RTN, 10 of 162 capped Nemotron samples and 40 of 233 capped Qwen3.5 samples are exact loops, against 12 of 36 and 53 of 169 under FP32.
We did not rerun the capped code samples at a larger budget.

\section{Teacher-forced KL misranks head groups in Nemotron-Nano-v2}
\label{sec:heads}

\begin{table}[t]
\centering
\small
\caption{Nemotron-Nano-9B-v2: \SR on subsets of heads, \RTN elsewhere. Octiles $q_0$ (slowest) to $q_7$ of the weights-only decay rate, 432 heads each. ``KL share'' is the share of the \bfs teacher-forced KL at positions beyond 4K tokens that is removed by keeping the octile in FP32 ($^\dagger$\,with \SR on all heads instead; $^\ddagger$\,sum of the two single-octile shares, not measured jointly). KL shares keep an octile in FP32, whereas the closed-loop arms apply \SR to it.}
\label{tab:heads}
\begin{tabular}{@{}lcccc@{}}
\toprule
\SR applied to & Heads & $\Flog$ [95\% CI] & Capped & KL share \\
\midrule
none (\bfs \RTN) & 0 & 0 & 31 & -- \\
all heads & 3456 & 1.05 [0.88, 1.27] & 0 & 94\%$^\dagger$ \\
$q_1{+}q_2$ ($r \in [0.57u, 4.1u)$) & 864 & 1.03 [0.86, 1.23] & 1 & 52\%$^\ddagger$ \\
$q_0{+}q_1$ (top-2 KL octiles) & 864 & 0.59 [0.44, 0.76] & 9 & 86\%$^\ddagger$ \\
$q_0{+}q_3$ (size-matched control) & 864 & $-$0.05 [$-$0.28, 0.14] & 34 & -- \\
random (211 in $q_1{+}q_2$) & 864 & 0.06 [$-$0.14, 0.24] & 34 & -- \\
all except $q_1{+}q_2$ & 2592 & 0.07 [$-$0.14, 0.26] & 26 & -- \\
\midrule
$q_0$ (slowest) & 432 & $-$0.12 [$-$0.37, 0.07] & 33 & 49\% \\
$q_1$ & 432 & 0.40 [0.24, 0.57] & 11 & 37\% \\
$q_2$ & 432 & 0.43 [0.25, 0.61] & 21 & 15\% \\
$q_3$ & 432 & $-$0.02 [$-$0.25, 0.16] & 37 & n/a \\
\bottomrule
\end{tabular}
\end{table}

A natural way to choose which state components need higher precision is a teacher-forced divergence: run fixed text under each policy and measure the KL to FP32.
\citet{lores} report that such a teacher-forced loss gap correlates with generation accuracy, \citet{kong} rank hybrid SSM-Transformer components for mixed-precision weight quantization by forward-pass KL, and \citet{lotfi} use token-level KL to locate where weight-quantized reasoning models start overthinking.
We test, in one model lineage, whether teacher-forced KL identifies the heads that drive the closed-loop length effect.

We ranked Nemotron-Nano-9B-v2's 3456 Mamba heads into octiles by weights-only decay rate.
On 120 reasoning traces from our math runs of this model (70 generated with \bfs \RTN state, 50 with FP32 state), we measured how much of the \bfs \RTN KL beyond 4K tokens disappears when one octile is kept in FP32.
The slowest octile $q_0$ accounts for the most (49\%), then $q_1$ (37\%) and $q_2$ (15\%).
The order is the same on the 50 FP32-generated traces alone (47\%, 34\%, 16\%), so it does not come from conditioning on \bfs text.
In closed-loop generation the effect does not follow this order (Table~\ref{tab:heads}).
\SR on $q_1{+}q_2$, 25\% of heads with per-step decay between $0.57u$ and $4.1u$, removes essentially all of the excess ($\Flog = 1.03$ [0.86, 1.23]).
\SR on $q_0$ has no detectable effect ($\Flog$ CI upper bound 0.07).
Both predictions were pre-registered, as was the follow-up with size-matched and single-octile arms.
The size-matched $q_0{+}q_3$ control removes nothing, so the $q_1{+}q_2$ result is not a head-count effect ($\Flog$ difference 1.08 [0.84, 1.40]).
At equal size, $q_2$ beats $q_0$ by 0.55 [0.34, 0.80] despite carrying a third of its KL share.
$q_1$ and $q_2$ contribute about equally (0.40 and 0.43); together they give 1.03, against a sum of 0.83 (exploratory: joint minus sum 0.19 [$-$0.08, 0.51]).

Two further pre-registered arms test allocation at a fixed budget of 864 heads.
\SR on the two octiles with the largest KL share ($q_0{+}q_1$) removes 59\% of the excess, less than the band (difference 0.44 [0.26, 0.64]).
We had predicted 0.29 by adding the single-octile values; the observed value is higher (exploratory: joint minus sum 0.30 [0.03, 0.65]), and $q_0$, which removes nothing alone, adds 0.19 [0.01, 0.39] on top of $q_1$.
Both octile pairs thus give more together than the sum of their parts, the first not significantly.
We read this as a lead for future work, not a finding: groups of heads may need \SR jointly to restore the dynamics.
\SR on 864 heads drawn uniformly at random, 211 of them in the band, removes no detectable excess (0.06 [$-$0.14, 0.24]; band minus random 0.97 [0.75, 1.26]).
We had predicted 0.26 from the band overlap.
So covering part of the band does not give a proportional part of the effect (exploratory): about a quarter of the band, spread over both octiles and mixed with 653 non-band heads, gives 0.06; one whole band octile gives 0.40--0.43.
These arms differ in composition, so we do not read them as a dose--response curve.

One teacher-forced behavioural signal does follow the closed-loop order; this analysis is exploratory.
On these traces, \bfs \RTN raises the teacher-forced expected rates of the four markers we tracked: ``Alternatively'' ($\times$1.23), ``Wait'' ($\times$1.13), ``Hmm'' ($\times$1.11) and ``But'' ($\times$1.10).
For ``Alternatively'', keeping an octile in FP32 removes 46\% ($q_2$), 30\% ($q_1$), 14\% ($q_3$) and $-2$\% ($q_0$) of the excess, the same order as the closed-loop $\Flog$.
The other three markers do not follow it: $q_0$ removes 26--31\% of their excess.
In closed-loop generation, of these four only ``Alternatively'' clearly rises under \bfs \RTN (1.14 to 1.39 per 1K output tokens, counted in the thinking text of items that both runs finish; Wilcoxon $p = 2\times10^{-10}$); ``Wait'', ``Hmm'' and ``But'' do not ($p \ge 0.12$).
``Let me check/verify'', which the teacher-forced analysis did not track, also rises (0.98 to 1.13, $p = 6\times10^{-5}$).
So among the tracked markers, the one that \bfs \RTN changes in generation is the one whose teacher-forced excess tracks the right heads, while the largest KL share comes from $q_0$, whose \SR alone has no detectable effect.
The evidence is 120 traces from one model.
\citet{lotfi} observed the same lexical markers (``wait'', ``but'', ``alternatively'') at high-KL positions under weight quantization.

The toy model explains the upper edge of the band (above about $4u$ \RTN is nearly unbiased) but not the lower edge.
$q_0$ heads freeze completely under \RTN in the toy, and on teacher-forced text they have quiet steps (data-dependent decay below $u$) as often as $q_1{+}q_2$ (82\% vs.\ 72\% of steps) and a similar number of stalled state elements per token.
Why their freezing does not change behaviour remains open.

\paragraph{Transfer.}
In the 12B parent of the same lineage, \SR on heads in the same absolute decay band removes most of the excess ($\Flog = 0.84$ [0.68, 1.03]; caps 2; pre-registered, pass).
In Nemotron-3 it does not: band-only \SR removes 19 of the 45 extra caps, the complementary heads 18, and all heads 40.
The effect there appears spread across heads (exploratory).
In Qwen3.5-9B (GDN) the same absolute band holds 174 of 768 heads, and all three pre-registered tests fail: band-only \SR removes 0.35 [$-$0.01, 0.64] of the excess, the other 594 heads 0.67 [0.36, 0.99], and the band does not beat 174 random heads (0.10; difference 0.24 [$-$0.10, 0.65]).
Here too the effect looks spread, with the share removed roughly following the share of heads with \SR (exploratory; all heads: 0.97).
The concentration in one decay band is a property of Nemotron-Nano-v2, not of recurrent hybrids in general.
This bears on decay-based precision allocation.
\citet{lores} keep the slowest-decaying half of heads at higher precision; in Nemotron-Nano-v2 that half includes our band.
A 25\% budget of the slowest heads ($q_0{+}q_1$), which here is also the set with the largest teacher-forced KL share, recovers 0.59 of the excess against 1.03 for the band, and the slowest octile alone recovers nothing.
Any such selection should be validated by closed-loop generation length.

\section{Recommendations for evaluation}
\label{sec:recs}

\begin{enumerate}
\item \textbf{Record the state dtype.} Report the resolved recurrent-state dtype and rounding mode for every run, not just the engine flag. Engine defaults vary by model and release~\citep{vllm63218}. Defaults also differ between engines: for a Qwen3.5 config without \texttt{mamba\_ssm\_dtype}, vLLM~0.33 uses the model dtype, usually \bfs~\citep{vllm033}, and SGLang~0.5.21 uses FP32~\citep{sgl0521}; a fine-tune that drops the field silently changes it.
\item \textbf{Report cap rate and output length with accuracy.} Here the precision change moved accuracy mainly through cap hits. Budget choice changes measured gaps~\citep{couplingtax,mindcap}, and Nemotron~3.5 caps the verbosity increase of its quantized checkpoint at 5\%~\citep[\S4]{n35}.
\item \textbf{Do not read ``both finished'' accuracy as ``no capability loss''.} That subset is selected and often near ceiling. Rerun capped items with a larger budget to separate longer reasoning from non-termination, and count exact loops.
\item \textbf{Validate state-precision changes with generation, not KL alone.} Teacher-forced KL ranked head groups differently from their closed-loop effect in our case.
\item \textbf{Prefer FP32 state, or 16-bit with \SR.} \SR fixed four of our five affected models but not Nemotron-3, so validate it per model. \bfs \SR is not yet available in SGLang~\citep{sgl26929,sgl27975}. It needs no extra memory and, with a hash generator, was no slower than \RTN on Qwen3.5-9B (\S\ref{sec:sr}).
\end{enumerate}

\section{Related work}
\label{sec:related}

\paragraph{Recurrent-state quantization.}
The closest work is NVIDIA's: FP16 \RTN SSM cache raises verbosity by up to 37\% and \SR fixes it~\citep{super}, with INT8 and FP8 results in \citet{ultra} and an FP16+\SR cache in a released runtime~\citep{n35}.
\citet{lores} show the same \RTN-versus-\SR length effect for INT8 on Qwen-3.5 and Nemotron-H.
ReplaySSM caches inputs instead of state and analyses accuracy on runs that both finish~\citep{replayssm}.
\citet{gdnlowbit} report longer outputs with FP16 state on Qwen3.5 GDN models and leave the mechanism open.
LeapQuant~\citep{leapquant} and DAMP~\citep{damp} report \bfs accuracy losses and propose better low-bit schemes.
\citet{dither} formalises write erasure under \RTN and measures its long-horizon effect with teacher-forced KL; \citet{preserving} show that lost small updates make state quantization errors coherent across steps.
We add the \bfs case on math and code, same-width \bfs \SR for Mamba-2 and GDN at \RTN decoding speed, a 128K budget analysis, and closed-loop head-localization tests in four models (positive in one lineage).

\paragraph{Length, budgets and numerics in evaluation.}
Weight quantization also lengthens chains of thought~\citep{lotfi}.
KL-based sensitivity is a common way to allocate precision in hybrid models~\citep{kong}; in our setting it ranks state heads differently from their effect on generation length.
\bfs nondeterminism alone can change response length by thousands of tokens~\citep{nondet}; our effects are paired, directional and large relative to the seed-to-seed variation we measured.
\citet{chatelain} vary only the rounding rule at a fixed format in a small GPT-2, as in our same-width \SR test; they study weights and activations under perplexity, not recurrent state in long generation.

\section{Limitations}
\label{sec:limits}

Prompts are competition math for all six models and competitive programming for two; all runs use one engine version on one GPU type.
Most arms have a single seed; Nemotron-Nano-9B-v2 has two seeds per main arm, Nemotron-3 two FP32 seeds and Qwen3.5-9B two \bfs \RTN runs.
The 128K analysis covers two models, with only 32 rerun items for Nemotron-Nano-9B-v2.
Where the FP32 request cap binds, concurrency differs between FP32 and \bfs arms; we matched it on Nemotron-3, and on Qwen3.5-9B the cap never binds.
\SR covers decode only; prefill still uses \RTN.
At matched batch size, \SR decodes at 96--101\% of same-day \bfs \RTN throughput on the Mamba-2 models, but the main-arm GDN kernel with Triton's default 10-round Philox~\citep{philox} decodes at 30--60\% (Qwen3.5-9B) and 36--85\% (Qwen3-Next), lower at larger batches.
In a kernel microbenchmark at Qwen3.5-9B shapes and batch 240 it takes $\times$20 the \RTN kernel time; 4-round Philox or the integer hash lowbias32~\citep{lowbias32} brings this to $\times$1.01 ($\times$1.68 and $\times$1.19 at batch 64, where the \bfs state fits in L2), and both pass the same bitwise and unbiasedness checks.
The hash variant shows no slowdown on Qwen3.5-9B, end to end or at matched batch size (\S\ref{sec:sr}); we did not rerun Qwen3-Next or the code benchmark with it.
Head localization holds in one lineage: band sufficiency replicates in the 12B parent ($\Flog = 0.84$), but the size-matched, random and single-octile controls were run on the 9B only. It fails to transfer to Nemotron-3, where \SR also leaves an unexplained bulk shift, and to Qwen3.5-9B.
The Falcon-H1R null is unexplained.
Our prior-art search covered arXiv and the SGLang and vLLM repositories only.

\bibliographystyle{tmlr}
\bibliography{refs}

\clearpage
\appendix
\section{Pre-registered tests}
\label{app:prereg}

We wrote each criterion below in our working notebook before the arm it concerns produced any data.
$\Flog$ uses the model's own FP32 and \bfs \RTN arms as references, with a 95\% bootstrap CI over items.
``Size'' is a geometric-mean length ratio with a Wilcoxon signed-rank test.
Table~\ref{tab:prereg} lists every pre-registered test, including those that failed and one whose analysis is not in the main text.

\begin{table}[!ht]
\centering
\footnotesize
\caption{Pre-registered tests and outcomes.}
\label{tab:prereg}
\begin{tabular}{@{}>{\raggedright\arraybackslash}p{0.25\linewidth}>{\raggedright\arraybackslash}p{0.34\linewidth}>{\raggedright\arraybackslash}p{0.23\linewidth}l@{}}
\toprule
Test & Criterion & Result & Verdict \\
\midrule
\multicolumn{4}{@{}l}{\emph{Head localization, Nemotron-Nano-9B-v2 (\S\ref{sec:heads})}}\\
\SR on $q_1{+}q_2$ & $\Flog > 0.5$, CI lower $> 0.25$; caps $\le 10$ & 1.03 [0.86, 1.23]; caps 1 & pass \\
\SR on $q_0$ & $\Flog < 0.33$ & $-0.12$ & pass \\
\SR on $q_0{+}q_3$ (size-matched) & $\Flog < 0.33$ & $-0.05$ & pass \\
$q_2$ vs.\ $q_0$ (equal size) & $\Flog$ difference CI lower $> 0$ & 0.55 [0.34, 0.80] & pass \\
\SR on $q_0{+}q_1$ (top-2 KL octiles) & $\Flog(q_1{+}q_2) - \Flog(q_0{+}q_1)$ CI lower $> 0$; fail if $\Flog(q_0{+}q_1) \ge 0.8$ & 0.44 [0.26, 0.64]; $\Flog$ 0.59 (predicted 0.29) & pass \\
\SR on 864 random heads & $\Flog(q_1{+}q_2) - \Flog(\text{random})$ CI lower $> 0$; fail if $\Flog(\text{random})$ CI includes 1 & 0.97 [0.75, 1.26]; $\Flog$ 0.06 (predicted 0.26) & pass \\
\midrule
\multicolumn{4}{@{}l}{\emph{Transfer of the band and of \SR}}\\
Falcon-H1R-7B, band \SR & only if \bfs inflates ($\times{>}1.05$, $p{<}0.01$) & no inflation ($\times$1.008) & untestable \\
Nemotron-Nano-12B-v2 (a) & \bfs inflates ($\times{>}1.05$, $p{<}0.01$) & $\times$1.188 & pass \\
\quad (b) full \SR, (c) band \SR & $\Flog > 0.5$, CI lower $> 0.25$ & 0.98; 0.84 [0.68, 1.03] & pass, pass \\
Nemotron-3 (a) & gate; no directional prediction & $\times$1.081 & gate met \\
\quad (b) full \SR, (c) band \SR & as 12B (b), (c) & 0.38 [$-$0.37, 1.00]; $-$0.27 & fail, fail \\
Nemotron-3, matched concurrency & $\times{\ge}1.05$ ($p{<}0.01$) and caps $\ge 59$ & $\times$1.098; caps 82 & pass \\
Qwen3.5-9B, GDN \SR & caps $\le 170$; $\Flog > 0.5$, CI lower $> 0.25$; secondary: McNemar \SR vs.\ \RTN $p{<}0.05$ & caps 111; 0.87 [0.58, 1.26]; 121 vs.\ 22 & pass \\
\quad hash generator (a) & $\Flog > 0.5$, CI lower $> 0.25$ & 0.97 [0.70, 1.37] & pass \\
\quad (b) throughput & tokens/s $\ge 0.9\times$ a same-day \RTN run & $\times$1.02 & pass \\
Qwen3-Next, GDN \SR & $\Flog > 0.5$, CI lower $> 0$; secondary: caps $\le 20$ & 0.99 [0.68, 1.41]; caps 15 & pass \\
Qwen3.5-9B, band \SR (a) & $\Flog > 0.5$, CI lower $> 0.25$ & 0.35 [$-$0.01, 0.64] & fail \\
\quad (b) other 594 heads & $\Flog < 0.33$ & 0.67 [0.36, 0.99] & fail \\
\quad (c) band vs.\ random & $\Flog$ difference CI lower $> 0$ & 0.24 [$-$0.10, 0.65] & fail \\
\midrule
\multicolumn{4}{@{}l}{\emph{Decoding, budget and benchmark}}\\
Greedy, Nemotron-Nano-9B-v2 & \bfs/FP32 size $> 1$, $p{<}0.01$ & $\times$1.211 & pass \\
Greedy, Qwen3.5-9B (secondary) & same & $\times$1.172 & pass \\
128K rerun, Qwen3.5-9B & relaxed accuracy \bfs $<$ FP32, McNemar $p{<}0.05$ & 86.6 vs.\ 93.6\% & pass \\
\quad & unfinished at 128K \bfs $>$ FP32, McNemar $p{<}0.05$ & 66 vs.\ 17 & pass \\
Cap simulation & FP32$-$\bfs accuracy gap at a 16K cap $>$ gap at 32K, on both Nemotron-Nano-9B-v2 and Qwen3.5-9B & Nemotron $+$2.2\,pp; Qwen3.5 $-$3.7\,pp & fail \\
LiveCodeBench, length (\S\ref{sec:lcb}) & \bfs/FP32 size CI lower $> 1$ & Nemotron $\times$1.37 [1.29, 1.46] (predicted 1.05--1.30); Qwen3.5 $\times$1.09 [1.05, 1.13] (predicted 1.10--1.40) & pass, pass \\
\quad \SR & $\Flog \ge 0.5$, CI lower $> 0$ & 0.95 [0.77, 1.15]; 0.73 [0.27, 1.33] & pass, pass \\
\quad caps (secondary) & caps \bfs $\ge$ FP32; sign test over problems & 36$\to$162 (65 vs.\ 2); 169$\to$233 (39 vs.\ 3) & pass, pass \\
\bottomrule
\end{tabular}
\end{table}

\clearpage
\section{Serving details}
\label{app:serving}

Table~\ref{tab:serving} lists the per-model settings; all arms use SGLang 0.5.21, a 40{,}960-token context and a 32{,}768-token output cap unless stated.
Every prompt (at most 1{,}762 tokens) fits in one 8{,}192-token prefill chunk, so prefill writes the cached state once per request, against up to 32{,}768 decode writes.
The running-request cap is set by the engine from the memory left for the recurrent-state cache, so \bfs arms admit about twice as many concurrent requests per rank as FP32 arms; \SR arms match \RTN arms exactly. On Qwen3.5-9B no arm exceeds 74 running requests per rank, below the FP32 cap of 122, so the cap never binds.
The Nemotron-9B, Qwen3.5-9B (main) and Qwen3-Next FP32 arms left \texttt{-{}-mamba-ssm-dtype} unset; SGLang then reads the model config and falls back to \texttt{float32}, and their logged state memory per cache slot is exactly twice that of the \bfs arms. If the config has no \texttt{mamba\_ssm\_dtype}, SGLang's resolver keeps its \texttt{float32} default~\citep{sgl0521}; deleting the field from the Qwen3.5-9B config confirms this.
Falcon-H1R runs with \texttt{-{}-cuda-graph-backend-prefill disabled} in all arms, to avoid SGLang issue~\#42774~\citep{sgl42774}.
The LiveCodeBench arms all use DP8 with the per-model sampling settings above; each returned all 350 samples, and the engine logged two request retractions in the Qwen3.5-9B \bfs arm and none in the other five.

\begin{table}[!ht]
\centering
\small
\caption{Serving settings. ``Running cap'' is the engine's per-rank maximum of concurrent requests, read from the server logs.}
\label{tab:serving}
\footnotesize
\begin{tabular}{@{}lccc@{}}
\toprule
Hugging Face repository & Parallelism & Running cap FP32 / \bfs & $T$ / top-$p$ / top-$k$ \\
\midrule
\texttt{nvidia/NVIDIA-Nemotron-Nano-9B-v2} & DP8 & 44 / 88 & 0.6 / 0.95 / -- \\
\texttt{nvidia/NVIDIA-Nemotron-Nano-12B-v2} & DP8 & 38 / 76 & 0.6 / 0.95 / -- \\
\texttt{nvidia/NVIDIA-Nemotron-3-Nano-30B-A3B-BF16} & DP8 & 43 / 86 & 1.0 / 1.0 / -- \\
\texttt{Qwen/Qwen3.5-9B} & DP6 / DP8$^*$ & 122 / 240 & 0.6 / 0.95 / 20 \\
\texttt{Qwen/Qwen3-Next-80B-A3B-Thinking} & TP4$\times$DP2 & 223 / 436 & 0.6 / 0.95 / 20 \\
\texttt{tiiuae/Falcon-H1R-7B} & DP8 & 47 / 94 & 0.6 / 0.95 / -- \\
\bottomrule
\end{tabular}
\\[2pt]
{\footnotesize $^*$The main FP32 arm used DP6 and the \bfs arms DP8; the 8-rank FP32 rerun (\S\ref{sec:setup}) logs the same per-rank cap of 122.}
\end{table}

The \SR store replaces the engine's FP32-to-\bfs conversion of each updated state value $h$:
\begin{verbatim}
  bits = bitcast_uint32(h)                         # FP32 value after the update
  bits = (bits + (rand32 & 0xFFFF)) & 0xFFFF0000   # SR
  # RTN (nearest-even) is equivalent to adding 0x7FFF + ((bits >> 16) & 1)
  store bitcast_float32(bits) as bf16              # exact: low 16 bits are zero
\end{verbatim}
\texttt{rand32} is a 32-bit draw from Triton's Philox generator, with a fresh seed at every CUDA-graph replay (\S\ref{sec:setup}).
The hash variant (\S\ref{sec:sr}) instead sets \texttt{rand32} to lowbias32(\texttt{i * 0x9E3779B1 + seed})~\citep{lowbias32}, where \texttt{i} is the element's offset in the state cache and \texttt{seed} is the same per-replay seed, a fresh 32-bit draw (\texttt{torch.randint(0, 2**32)}) at every decode step.
Per-step unbiasedness does not imply independence across steps, so we checked that separately in a CPU emulation of the kernel's formula ($2^{20}$ elements, 256 steps, criteria set before running): the correlation between an element's draws at lags of 1 to 4 steps is at most $4.5\times10^{-5}$ (limit $2.5\times10^{-4}$), the variance of each element's 256-step sum is 1.002 times its independent-draw value, and in the toy model of Figure~\ref{fig:toy} the effective decay with hash bits is within 0.0012 of that with independent random bits for every $r$ from $0.25u$ to $16u$.

\end{document}